\documentclass[letterpaper, 10 pt, conference]{ieeeconf}  

\IEEEoverridecommandlockouts                              

\usepackage{booktabs}
\usepackage{cite}
\usepackage{float}
\usepackage{graphicx}
\usepackage{amssymb}
\usepackage{times}
\everymath{\displaystyle}
\usepackage{booktabs}

\usepackage{amsthm}
\usepackage{xfrac}
\usepackage{balance}
\usepackage{color}
\usepackage{mathtools}
\usepackage{tabularx}
\usepackage{bm}
\usepackage{diagbox}
\usepackage{float}
\usepackage{epstopdf}
\usepackage{url}
\usepackage{multirow}
\usepackage{tikz}
\usepackage[linkcolor=black,citecolor=black,urlcolor=black,colorlinks=true]{hyperref}
\let\labelindent\relax
\usepackage{enumitem}
\usepackage[ruled,vlined,linesnumbered]{algorithm2e}
\usepackage[T1]{fontenc}
\usepackage{aecompl}

\title{\LARGE \bf
LIKO: LiDAR, Inertial, and Kinematic Odometry for Bipedal Robots
}

\author{Qingrui Zhao$^{1}$, Mingyuan Li$^{1}$, Yongliang Shi$^{2}$, Xuechao Chen$^{1}$, Zhangguo Yu$^{1}$, Lianqiang Han$^{1}$,
\\Zhenyuan Fu$^{1}$, Jintao Zhang$^{1}$, Chao Li$^{1}$, Yuanxi Zhang$^{1}$, Qiang Huang$^{1}$ 
\thanks{*This work was supported in part by the National Natural Science Foundation of China under Grant 62073041, Grant 62088101, and the Beijing Municipal Science and Technology Project under Grant Z231100007123006.}
\thanks{$^{1}$School of Mechatronical Engineering, Beijing Institute of Technology (BIT), Beijing, China. Contact: 
        {\tt\small chenxuechao@bit.edu.cn}}%
\thanks{$^{2}$Qiyuan Lab}%
}

\begin{document}

\maketitle

\begin{abstract}

High-frequency and accurate state estimation is crucial for biped robots. This paper presents a tightly-coupled LiDAR-Inertial-Kinematic Odometry (LIKO) for biped robot state estimation based on an iterated extended Kalman filter. Beyond state estimation, the foot contact position is also modeled and estimated. This allows for both position and velocity updates from kinematic measurement. Additionally, the use of kinematic measurement results in an increased output state frequency of about 1kHz. This ensures temporal continuity of the estimated state and makes it practical for control purposes of biped robots. We also announce a biped robot dataset consisting of LiDAR, inertial measurement unit (IMU), joint encoders, force/torque (F/T) sensors, and motion capture ground truth to evaluate the proposed method. The dataset is collected during robot locomotion, and our approach reached the best quantitative result among other LIO-based methods and biped robot state estimation algorithms. The dataset and source code will be available at \href{https://github.com/Mr-Zqr/LIKO}{\tt\small https://github.com/Mr-Zqr/LIKO}.

\end{abstract}

\section{Introduction}

Biped robots are developing rapidly these days, gradually stepping out of laboratories and into the real world. In contrast to wheeled vehicles or quadrotors that maintain continuous ground contact or experience smooth acceleration patterns, legged robots, especially bipedal ones, encounter intermittent ground contact during locomotion \cite{wisth2022vilens}. This introduces additional sensor noise to IMU and joint encoders. Therefore, smooth and precise linear velocity estimation is imperative for stable closed-loop control in bipedal robots \cite{liu2022design}. Furthermore, bipedal robots are inherently more prone to instability and falls compared to their quadrupedal counterparts. Consequently, a high-frequency (over 1kHz) controller is essential to ensure stability in balance and gait control \cite{ahn2023development}, \cite{wang2022design}. Similarly, the state estimator should also be high-frequency (250Hz to 1kHz) and globally accurate to enable the formation of a stable feedback control system \cite{camurri2020pronto}.

Existing state estimation algorithms for legged robots mainly use proprioceptive sensors, such as IMUs, joint encoders, and contact sensors. These sensor measurements are fused using filtering-based methods, such as unscented Kalman filter \cite{bloesch2013stateslippery}, extended Kalman filter \cite{bloesch2017state}, and invariant extended Kalman filter (InEKF) \cite{hartley2020contact}. These methods enable real-time and high-frequency state estimation. However, according to \cite{bloesch2013stateslippery}, the robot's global position and yaw angle cannot be observed using only proprioceptive sensors, resulting in drift. 

To mitigate the position drift and make full use of the environmental information, some legged robots are equipped with exteroceptive sensors, namely stereo cameras and LiDARs. The introduction of these sensors also comes with a large computation load. To manage this computational load without sacrificing real-time performance, the loosely coupled approach has been employed \cite{camurri2020pronto}. However, this method solely relies on the estimated state from each sensor and lacks the capability for online estimation of sensor parameters, leading to suboptimal state estimation. Recently, a tightly coupled method has also been presented \cite{wisth2022vilens}, which tightly fuses visual, inertial, LiDAR, and kinematic measurements through factor graphs. However, it is designed for quadruped robots and does not include foothold estimation. 

\begin{figure}[t]
    \centering
    \includegraphics[width=0.96\linewidth]{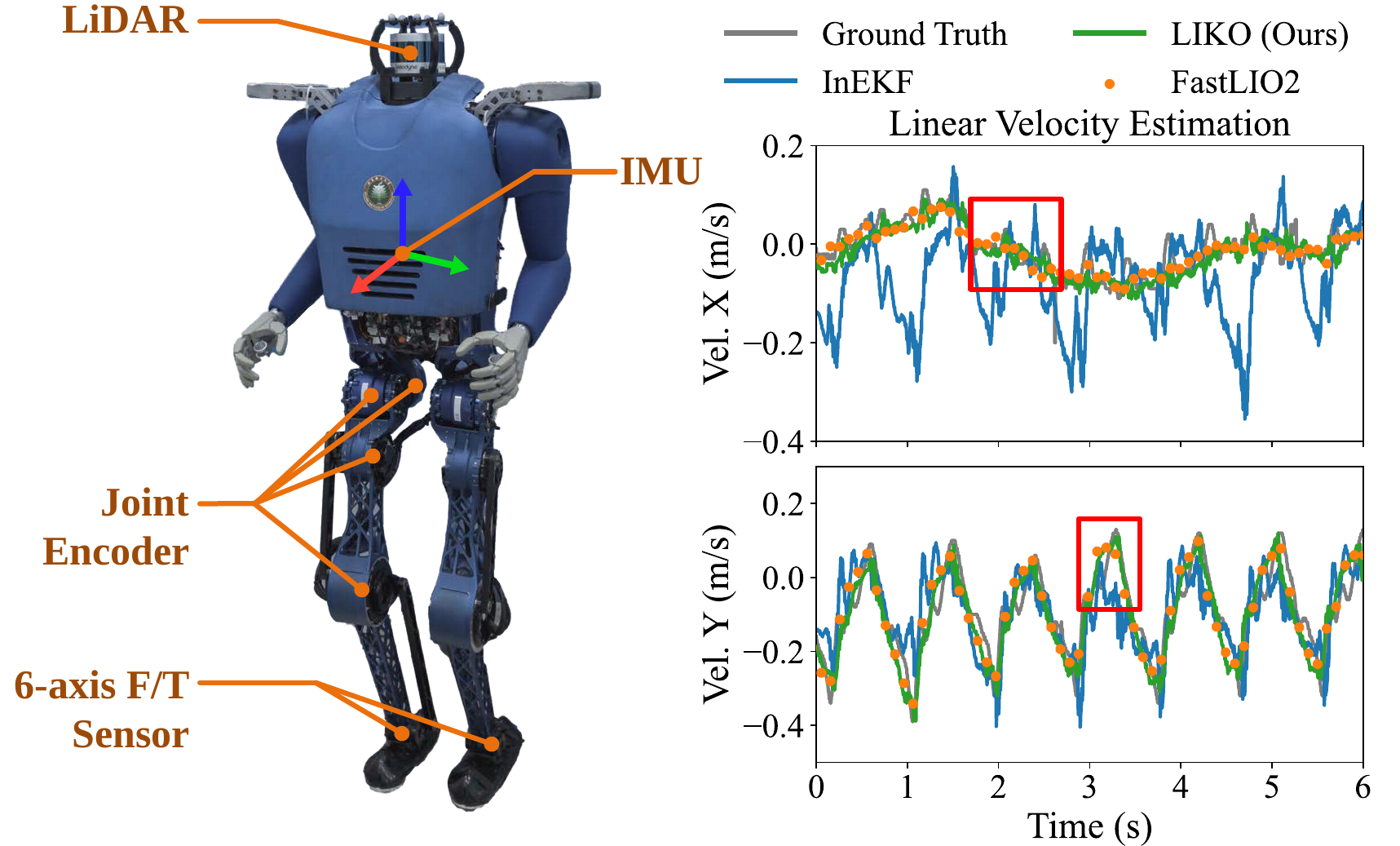}
    \caption{\textit{Left:} The proposed LIKO state estimation system is tested on a BHR-B3 biped robot. The robot's body coordinate is aligned with the IMU coordinate. \textit{Right:} Linear velocity estimation comparison between proposed LIKO (green), Inertial Kinematic odometry Contact-aided InEKF \cite{hartley2020contact} (blue), and LiDAR-Inertial odometry (LIO) FastLIO2 \cite{xu2022fast} (orange) against ground truth (gray). The velocity estimated by FastLIO2 is plotted in scatter form since its update frequency is at 10Hz with Velodyne LiDAR, while other methods operate at 1kHz. The red boxes highlighted where the proposed method exhibits more accurate velocity estimation than others. }
    \label{fig:teaser-fig}
\end{figure}

To address the above limitations, we present LiDAR-Inertial-Kinematic Odometry (LIKO), a tightly-coupled high-frequency and accurate odometry for biped robot state estimation. As shown in Fig. \ref{fig:teaser-fig}, our method achieved high-frequency, smooth, and accurate velocity estimation against other state-of-the-art state estimation algorithms for biped robots and LiDAR-inertial odometry.  More specifically, our contributions are as follows: 
\begin{itemize}
    \item LIKO: a tightly-coupled LiDAR-Inertial-Kinematic Odometry to achieve high-frequency and accurate state estimation for biped robots. 
    \item Online estimation of the foot contact position, leveraging both position and velocity from the leg odometry as the measurement update of the state.    
    \item Datasets containing multiple motion patterns of a biped robot and corresponding motion-captured ground truth are collected and announced to test LIKO.
    \item Experimental validations were conducted on hardware and provided datasets to assess the performance of the proposed algorithm. The results demonstrated a 14\% improvement in accuracy compared to the current state-of-the-art methods.
\end{itemize}

\section{Related Works}
\subsection{Biped Robot State Estimation}
Many legged robot state estimation algorithms focus on fusing IMU and kinematic measurements using filter-based methods. Kinematic measurements, commonly referred to as leg odometry \cite{camurri2020pronto}, leverage joint encoders and contact sensors, typically F/T sensors, in conjunction with robot forward kinematics to derive position and velocity information under the non-slipping assumption \cite{bloesch2013statefusion}, \cite{bledt2018cheetah}. Bloesch \textit{et al.} \cite{bloesch2013stateslippery} introduced the application of an Unscented Kalman filter to robot state estimation. Hartley \textit{et al.} \cite{hartley2020contact} proved that the system dynamics of bipedal robots satisfy the group affine property, leading to the introduction of a contact-aided InEKF for better convergence in orientation estimation. However, the leg odometry in this work is limited to using only position measurements. Xavier \textit{et al.} \cite{xavier2023multi} modeled the contact velocity by attaching additional IMUs to robot feet, which released the non-slipping assumption and achieved better performance in long-term experiments. Nonetheless, as demonstrated in \cite{bloesch2013stateslippery}, solely relying on proprioceptive sensors results in the non-observability of global position and yaw angle, leading to position drift.

To address the position drift problem and make full use of the environmental information, some legged robots also use exteroceptive sensors, such as stereo cameras and LiDARs \cite{wisth2019robust, fallon2014drift, yang2023cerberus, teng2021legged}. Camurri \textit{et al.} \cite{camurri2020pronto} used an extended Kalman filter (EKF) to loosely-couple estimations from visual, inertial, LiDAR, and leg odometry. Wisth \textit{et al.} \cite{wisth2022vilens} proposed VILENS, which tightly coupled all sensor modalities mentioned above with a factor graph. However, since the factor graph optimizes a large number of states at a time, this method is computationally intensive by nature. Fallon \textit{et al.} \cite{fallon2014drift}  used EKF in conjunction with IMU, joint encoder, and LiDAR to estimate the state of the Atlas robot.  However, it did not include contact position estimation. The leg odometry only provides linear velocity measurement by taking the derivative of forward kinematics.

\subsection{LiDAR-Inertial Odometry}
LiDAR odometry entails the estimation of a robot's pose and position through LiDAR sensors. This is achieved by determining the relative transformation between two consecutive sensor frames with scan-matching techniques such as the Iterative Closest Point (ICP) \cite{besl1992method} or Generalized-ICP (GICP) \cite{segal2009generalized}. Recently, some feature-based approaches \cite{zhang2014loam, shan2018lego, pan2021mulls} have been reported, which selectively use the most salient data points to perform the scan-matching algorithm. However, the presence of noise stemming from LiDAR's rotation mechanism and sensor movements during scanning often introduces distortions in the acquired point clouds. Thus, relying solely on LiDAR for state estimation is not ideal, as registering distorted point clouds can lead to cumulative errors, ultimately resulting in odometry drift.

IMU is commonly employed to address such challenges, forming LiDAR-Inertial odometry (LIO). Loosely-coupled LIO typically processes LiDAR and IMU measurements separately and fuses them using a Kalman filter. In LeGO-LOAM \cite{shan2018lego}, IMU measurements are utilized for de-skewing lidar scans and providing motion prior to scan-matching. Conversely, tightly-coupled LIO systems generally utilize raw LiDAR data in conjunction with IMU data. FAST-LIO series \cite{xu2022fast}, \cite{xu2021fast} employed an Iterated Extended Kalman Filter (IEKF) to fuse LiDAR and IMU measurements. Additionally, they introduced a back-propagation approach to estimate the relative motion between the sample time of each point and the scan end time. Recently, DLIO \cite{chen2023direct} introduced a coarse-to-fine motion correction method and a hierarchical geometric observer \cite{lopez2023contracting} for data fusion.

\section{Methodology} \label{sec: method}
\begin{figure*}
\centering
\includegraphics[width=0.96\linewidth]{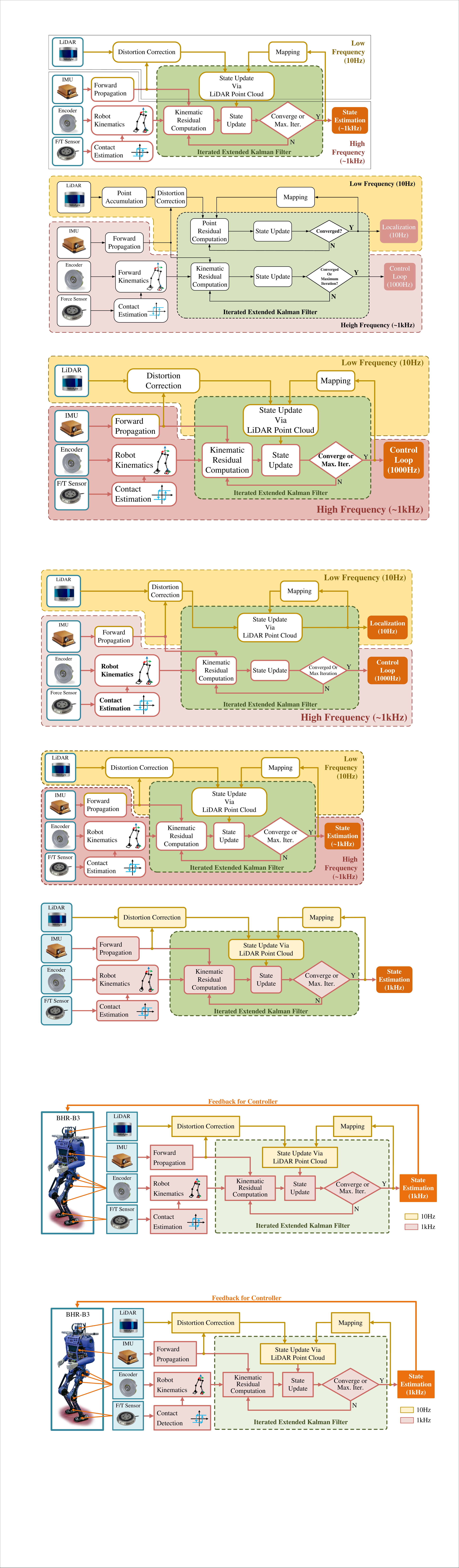}
\caption{System overview of LIKO. The system receives input from a LiDAR, an IMU, joint encoders, and F/T sensors. The IMU is used for state propagation, while LiDAR, joint encoder, and F/T sensors generate three different types of state measurement. The formulation of the iterated Kalman filter is discussed in Section \ref{sec: method}.}
\label{fig:system overview}
\end{figure*}
We begin by introducing the relevant notation and conventions. In general, we denote scalars and frame abbreviations in lowercase italics, matrices in uppercase Roman bold, and vectors in lowercase Roman bold. The operator $\lfloor\mathbf{a}\rfloor_{\wedge}$ is used to convert a vector $\mathbf{a}$ into its corresponding skew-symmetric matrix. We also use $\mathbf{x}$, $\check{\mathbf{x}}$, and $\hat{\mathbf{x}}$ to represent the ground-truth, propagated, and updated values of $\mathbf{x}$ respectively. The notation $\hat{\mathbf{x}}^{j}$ denotes the $j$-th update of $\mathbf{x}$ in the IEKF. $\boxplus$ and $\boxminus$ are two encapsulation operators that represent a bijective mapping from a local neighborhood on $\mathcal{M}$ to its tangent space $\mathbb{R}^n$ \cite{xu2021fast}. Besides, $\tilde{\mathbf{x}}$ represents the discrepancy between $\mathbf{x}$ and $\hat{\mathbf{x}}$, and ${\left( \cdot \right)}^m$ stands for sensor measurements.

\subsection{System Overview}
The overview of the presented state estimation system is shown in Fig. \ref{fig:system overview}. The system receives input from a LiDAR, an IMU, joint encoders, and F/T sensors. Acceleration and angular velocity measurements from the IMU are first integrated to produce state propagation. The propagated states are then updated in IEKF using Robot Kinematics and LiDAR measurements. In Robot Kinematics measurement, F/T sensors determine which foot is in contact with the ground. Thereafter, the robot's position and linear velocity are computed, integrating data from joint encoders, foothold estimation, and forward kinematic introduced in Section \ref{kinematic measurement}. Meanwhile, LiDAR points of a scan are accumulated and fed into the distortion correction module with IMU measurements for motion compensation. The rectified point cloud is then used as a LiDAR measurement. The IEKF-updated state is then used for the control module and point cloud registration. 

\subsection{State Representation}
The robot state $\mathbf{x}$ is defined as follows: 
\begin{equation}~\label{e: state vector}
\begin{aligned}
    \mathbf{x} &\triangleq 
    \begin{bmatrix}
    { }^G \mathbf{R}_I & { }^G \mathbf{p}_I^T & { }^G \mathbf{v}_I^T & {\mathbf{b}^{\boldsymbol{\omega}}}^T & {\mathbf{b}^{\mathbf{a}}}^T & {}^G\mathbf{p}_{c}^T &{}^G \mathbf{g}^T
    \end{bmatrix}^T \\
    \mathcal{M} &= SO(3)\times \mathbb{R}^{18}, \quad \mathbf{x}  \in \mathcal{M}
\end{aligned}
\end{equation}

where ${}^G \mathbf{R}_I \in SO(3)$, ${}^G \mathbf{p}_I \in \mathbb{R}^3$, ${}^G \mathbf{v}_I \in \mathbb{R}^3$ are orientation, position, and velocity of the IMU with respect to the global frame, respectively. $\mathbf{b}^{\boldsymbol{\omega}}, \! \mathbf{b}^{\boldsymbol{a}} \in \mathbb{R}^3$ are random walk biases of the IMU gyroscope and accelerometer; ${}^G\mathbf{p}_c \in \mathbb{R}^3$ is the contact position of the robot in the global frame, and ${}^G\mathbf{g} \in \mathbb{R}^3$ is the gravity vector in the global frame. 

The IMU measurements are modeled as real values corrupted by additive Gaussian white noise ($\mathbf{w}^{\mathbf{\boldsymbol{\omega}}},\mathbf{w}^\mathbf{a}$) and random walk bias ($\mathbf{b}^\mathbf{\boldsymbol\omega}, \mathbf{b}^\mathbf{a}$): 
\begin{equation} ~\label{e:imu model}
\begin{aligned}
    {\boldsymbol \omega}^m =\boldsymbol \omega + \mathbf{w}^\mathbf{\boldsymbol\omega} + \mathbf{b}^\mathbf{\boldsymbol\omega} , \quad
    {\mathbf{a}}^m = \mathbf{a} + \mathbf{w}^\mathbf{a} + \mathbf{b}^\mathbf{a}
\end{aligned}
\end{equation}

\subsection{Propagation Model}
The kinematic model of the state $\mathbf{x}$ can be written as: 
\begin{equation} ~\label{e: propagation model continus}
    \begin{aligned}
        { }^G \dot{\mathbf{R}}_I^T & ={ }^G \mathbf{R}_I\left\lfloor{\boldsymbol \omega}^{m}-\mathbf{b}^{\boldsymbol{\omega}}-\mathbf{w}^{\boldsymbol{\omega}}\right\rfloor_\wedge \\
        {}^G \dot{\mathbf{p}}_I &= {}^G\mathbf{v}_I\\
        { }^G \dot{\mathbf{v}}_I&={ }^G \mathbf{R}_I\left({\mathbf{a}^{m}}-\mathbf{b}^{\mathbf{a}}-\mathbf{w}^{\mathbf{a}}\right)+{ }^G \mathbf{g} \\
        \dot{\mathbf{b}}^{\boldsymbol{\omega}}& =\mathbf{w}^{\mathbf{b} \boldsymbol{\omega}}, \quad 
        \dot{\mathbf{b}}^{\mathrm{a}}=\mathbf{w}^{\mathbf{ba}}\\
        {}^G\dot{\mathbf{p}}_c &= {}^G\mathbf{R}_I\cdot \mathbf{fko}({\mathbf{q}})\cdot\mathbf{w}^c, \quad
        { }^G \dot{\mathbf{g}}=\mathbf{0} 
    \end{aligned}
\end{equation}

where $\mathbf{fko}(\mathbf{q})$ is the orientation of the contact frame with respect to the IMU frame computed through encoder measurements, ${\mathbf{q}}\in \mathbb{R}^M$, and the forward kinematics. Besides, we relax the static contact assumption by adding a contact noise $\mathbf{w}^c$. The continuous kinematic model in (\ref{e: propagation model continus}) can be discretized at each imu measurement $t$, with period $\Delta t$: 
\begin{equation} ~\label{e: imu model discretized}
\mathbf{x}_{t+1}=\mathbf{x}_t \boxplus\left(\Delta t \mathbf{f}\left(\mathbf{x}_t, \mathbf{u}_t, \mathbf{w}_t\right)\right)
\end{equation}
the function $\mathbf{f}$, input $\mathbf{u}$, and noise $\mathbf{w}$ are defined as below:
\begin{equation}
    \begin{aligned}
        \mathbf{f}\left(\mathbf{x}_t, \mathbf{u}_t, \mathbf{w}_t\right) & =
        \begin{bmatrix}
            \boldsymbol{\omega}^m_t-\mathbf{b}^{\boldsymbol{\omega}}_t-\mathbf{w}^{\boldsymbol{\omega}}_t \\
            { }^G \mathbf{v}_{I_t} \\
            { }^G \mathbf{R}_{I_t}\left({\mathbf{a}}^m_t-\mathbf{b}^{\mathbf{a}}_t-\mathbf{w}^{\mathbf{a}}_t\right)+{ }^G \mathbf{g}_t \\
            \mathbf{w}^{\mathbf{b} \boldsymbol{\omega}}_t \\
            \mathbf{w}^{\mathbf{b} \mathbf{a}}_t \\
            {}^G\mathbf{R}_{I_t}\cdot \mathbf{fko}({\mathbf{q}}_t)\cdot\mathbf{w}^c\\
            \mathbf{0}_{3 \times 1}
        \end{bmatrix} \\
        \mathbf{u} &\triangleq \begin{bmatrix}
            {\boldsymbol{\omega}^m}^T & {\mathbf{a}^m}^T
        \end{bmatrix}^T \\
    \mathbf{w} &\triangleq \begin{bmatrix}
        {\mathbf{w}^{\boldsymbol{\omega}}}^T & \!{\mathbf{w}^{\mathbf{a}}}^T & \!{\mathbf{w}^{\mathbf{b} \boldsymbol{\omega}}}^T &\! {\mathbf{w}^{\mathbf{b a}}}^T &{\mathbf{w}^c}^T
    \end{bmatrix}^T
    \end{aligned}
\end{equation}

The forward propagation of the state $\mathbf{x}_t$ follows (\ref{e: imu model discretized}) by setting $\mathbf{w}_t = \boldsymbol{0}$:
\begin{equation}
    \check{\mathbf{x}}_{t}=\hat{\mathbf{x}}_{t-1} \boxplus\left(\Delta t \mathbf{f}\left(\hat{\mathbf{x}}_{t-1},\mathbf{u}_t, \mathbf{0}\right)\right)
\end{equation}
and the covariance propagation: 
\begin{equation}
    \check{\mathbf{P}}_{t}=\mathbf{F}_{\tilde{\mathrm{x}}} \hat{\mathbf{P}}_{t-1} \mathbf{F}_{\tilde{\mathbf{x}}}^T+\mathbf{F}_{\mathrm{w}} \mathbf{Q} \mathbf{F}_{\mathrm{w}}^T
\end{equation}
where $\mathbf{F}_{\tilde{\mathbf{x}}}$ and $\mathbf{F}_{\mathbf{w}}$ are partial differentiation of $\tilde{\mathbf{x}}_{t}$ with respect to $\tilde{\mathbf{x}}_{t-1}$ and $\mathbf{w}_{t-1}$ separately. 

Notice that $\check{\mathbf{P}}_t$ represents the covariance of the following error state: 
\begin{equation}
    \mathbf{x}_t \boxminus \check{\mathbf{x}}_t = \hat{\mathbf{x}}_t^j \boxminus \check{\mathbf{x}}_t + \mathbf{J} \tilde{\mathbf{x}}_t \sim \mathcal{N}\left(\mathbf{0}, \check{\mathbf{P}}_t\right)
\end{equation}
where the right superscript $j$ stands for the $j$-th update in the iterated Kalman filter, $\mathbf{J}$ is the partial differentiation of $\mathbf{x}_t \boxminus \hat{\mathbf{x}}_t $ with respect to $\tilde{\mathbf{x}}_t$ evaluated at zero: 
\begin{equation}
\begin{aligned}
        \mathbf{J} &= 
    \begin{bmatrix}
        \mathbf{A}\left({}^G\hat{\mathbf{R}}_I^j \boxminus {}^G\check{\mathbf{R}}_I\right)^{-T} & \mathbf{0}_{3\times18} \\
        \mathbf{0}_{18\times3} & \mathbf{I}_{18\times18}
    \end{bmatrix} \\
    \mathbf{A}(\mathbf{u})^{-1}&=\mathbf{I}-\frac{\lfloor\mathbf{u}\rfloor_{\wedge}}{2}+\left(1-\frac{\|\mathbf{u}\|}{2} \cot \left(\frac{\|\mathbf{u}\|}{2}\right)\right) \frac{\lfloor\mathbf{u}\rfloor_{\wedge}}{\|\mathbf{u}\|}
\end{aligned}
\end{equation}

\subsection{LiDAR Measurement}
We model the LiDAR measurements the same way as in \cite{xu2021fast}. The $j$-th point in scan $k$ at the LiDAR frame is noted as ${}^L p_j$. After distortion correction using the IMU back-propagation, all feature points in scan $t_k$ are mapped to the scan end time $t_k$. Meanwhile, each feature point corresponds to a plane feature in the global map. The residual of LiDAR measurement is defined as: 
\begin{equation} ~\label{e: lidar residual}
\begin{aligned}
   \mathbf{0}&= \mathbf{h}_j\left(\mathbf{x}_t,{ }^L \mathbf{n}_j\right)\\ &=\mathbf{u}_j^{T}\left({ }^G \mathbf{T}_{I_k}{ }^I \mathbf{T}_L \left({}^L\mathbf{p}_j+{ }^L \mathbf{n}_j\right)-{ }^G \mathbf{q}_j\right) \\
\end{aligned}
\end{equation}

where $\mathbf{u}_j$ is the normal vector of the corresponding plane, ${}^G\mathbf{q}_j$ is the point on the corresponding plane, ${}^G\mathbf{T}_{I_k}$ is the relative pose between world frame and local frame, ${}^I\mathbf{T}_L$ is extrinsic, and ${}^L\mathbf{n}_j$ is the LiDAR ranging and beam-directing noise. 

Moreover, approximating the measurement equation (\ref{e: lidar residual}) by its first-order approximation at $\hat{\mathbf{x}}_t$ leads to: 

\begin{equation} ~\label{e: lidar first taylor}
    \begin{aligned}
\mathbf{0} & =\mathbf{h}_j\left(\mathbf{x}_t,{ }^L \mathbf{n}_j\right) \\
&\simeq \mathbf{h}_j\left(\check{\mathbf{x}}_t, \mathbf{0}\right)+\mathbf{H}_j \tilde{\mathbf{x}}_t+\mathbf{r}_j \\
\end{aligned}
\end{equation}
where $\tilde{\mathbf{x}}_t = \mathbf{x}_t\boxminus \hat{\mathbf{x}}_t$ is the state error, $\mathbf{r}_j \sim \mathcal{N}\left(\mathbf{0}, \mathbf{R}_j\right)$ is the noise of Gaussian distribution due to the raw LiDAR measurement ${}^L\mathbf{n}_j$, and $\mathbf{H}_j$ is the Jacobian matrix of the measurement model with respect to $\tilde{\mathbf{x}}_t$, evaluated at zero, which is given as: 
\begin{equation} 
    \begin{aligned}
        \mathbf{H}_j & = \left.\frac{\partial \mathbf{h}_j(\check{\mathbf{x}}_t\boxplus\tilde{\mathbf{x}}_t, \mathbf{0})}{\partial \tilde{\mathbf{x}}_t}\right|_{\tilde{\mathbf{x}}_t=\mathbf{0}} \\
        &= \mathbf{u}_j^T
        \begin{bmatrix}
            -{}^G\check{\mathbf{R}}_I \left\lfloor{}^I\mathbf{R}_L {}^L\mathbf{p}_j+{}^I\mathbf{t}_L \right\rfloor_\wedge & \mathbf{I}_{3\times 3} & \mathbf{0}_{3\times 12}
        \end{bmatrix}
    \end{aligned}
\end{equation}

\subsection{Kinematic Measurement} \label{kinematic measurement}
\subsubsection{{Contact Classification}}

The biped robot platform used in this work is equipped with F/T sensors at both feet. The contact state can be inferred by thresholding the measured normal force at 250 N. Additionally, we applied a Schmitt trigger to filter contact states to deal with noisy force sensor measurements.

\subsubsection{{Velocity Measurement}}

The position of the foot in contact with the ground can be written as:
\begin{equation} ~\label{e: foot position}
    {}^G\mathbf{p}_{foot} = {}^G\mathbf{R}_I\cdot\mathbf{fk}\left(\mathbf{q}\right)+{}^G\mathbf{p}_I
\end{equation}
where $\mathbf{q}$ are the joint positions and $\mathbf{fk}(\mathbf{q})$ is the forward kinematics function.

The linear velocity of the robot's floating base at time $t$ expressed in the global frame can be computed by taking the derivative of (\ref{e: foot position}) and assuming that the foot velocity is zero: 
\begin{equation} ~\label{e: base velocity}
    {}^G{\mathbf{v}}_{I} = -{}^G\mathbf{R}_I\left(\mathbf{J}\left(\mathbf{q}\right)\dot{\mathbf{q}}+\boldsymbol{\omega}_t\times\mathbf{fk}\left(\mathbf{q}\right)\right)
\end{equation}
where $\mathbf{J}(\mathbf{q})$ is the corresponding kinematic Jacobian. Both the joint positions and velocities are measured from encoders and corrupted by additive zero-mean Gaussian noise $\mathbf{w}^{\mathbf{q}}$ and $\mathbf{w}^{\dot{\mathbf{q}}}$: 
\begin{equation} ~\label{e: joint position measurement}
        {\mathbf{q}}^m = \mathbf{q} + \mathbf{w}^{\mathbf{q}} , \quad
    {\dot{\mathbf{q}}}^m = \dot{\mathbf{q}}+\mathbf{w}^{\dot{\mathbf{q}}}
\end{equation}

After substituting (\ref{e: joint position measurement})  into (\ref{e: base velocity}), we can formulate the measurement of base velocity: 
\begin{equation} ~\label{e: real velocity measurement}
\begin{aligned}
{}^G{\mathbf{v}_{I}}^{m} &= -{}^G\mathbf{R}_I \Bigg(\mathbf{J}\left({\mathbf{q}}^m-\mathbf{w}^{\mathbf{q}}\right)\cdot\left({\dot{\mathbf{q}}^m}-\mathbf{w}^{\dot{\mathbf{q}}}\right)\Bigg.+ \\
&\Bigg. ({\boldsymbol{\omega}^m}_t-\mathbf{b}_{\boldsymbol{\omega}})\times \mathbf{fk}\left({\mathbf{q}}^m-\mathbf{w}^{\mathbf{q}}\right)\Bigg)
\end{aligned}
\end{equation}
The velocity residual $\mathbf{r}_{cv}\left(\check{\mathbf{x}}_t,\mathbf{u}_t,\mathbf{q}, \dot{\mathbf{q}}\right)$ is defined as: 
\begin{equation} ~\label{e: velocity residial}
   \mathbf{r}_{cv}\left(\check{\mathbf{x}}_t,\mathbf{u}_t,\mathbf{q}, \dot{\mathbf{q}}\right) = \mathbf{z}_{cv} = {}^G\check{\mathbf{v}}_I - {}^G{\mathbf{v}_{I}}^{m}
\end{equation}
where ${}^G\check{\mathbf{v}}_I$ is the propagated base velocity from IMU propagation process. 

Then, we obtain the first-order Taylor expansion of the true zero residual $\mathbf{r}_{cv}\left({\mathbf{x}}_t,\mathbf{u}_t,\mathbf{q}, \dot{\mathbf{q}}\right)$ as: 
\begin{equation} ~\label{e: kinematic first order taylor}
\begin{aligned}
\mathbf{0} &= \mathbf{h}_{cv}\left({\mathbf{x}}_t,\mathbf{u}_t,\mathbf{w}^{\mathbf{q}}, \mathbf{w}^{\dot{\mathbf{q}}}\right) \\
&\simeq\mathbf{h}_{cv}\left({\check{\mathbf{x}}}_t,\mathbf{u}_t,\mathbf{0}, \mathbf{0}\right) + \mathbf{H}_{cv }\tilde{\mathbf{x}}_t+\mathbf{w}_{cv} \\
&= \mathbf{z}_{cv} + \mathbf{H}_{cv }\tilde{\mathbf{x}}_t+\mathbf{w}_{cv}
\end{aligned}
\end{equation}
where $\mathbf{w}_{cv} \sim \mathcal{N}(\mathbf{0}, \mathbf{\Sigma}_{cv})$ and 
\begin{equation} ~\label{e: velocity measurement H}
\begin{aligned}
\mathbf{H}_{cv} &=
\left.\frac{\partial\mathbf{r}\left(\check{\mathbf{x}}_t\boxplus\tilde{\mathbf{x}}_t,\mathbf{u}_t,\mathbf{q}, \dot{\mathbf{q}}\right)}{\partial \tilde{\mathbf{x}}_t}\right|_{\tilde{\mathbf{x}}_t=\mathbf{0}}\\
&= 
\begin{bmatrix}
    \mathbf{H}_{cv1} & \mathbf{0}_{3\times 3} & \mathbf{I}_{3} & {}^G\check{\mathbf{R}}_I\left\lfloor\mathbf{fk}\left(\mathbf{q}\right)\right\rfloor_{\wedge} & \mathbf{0}_{3\times 9}
\end{bmatrix} \\
\mathbf{H}_{cv1} 
&= {}^G\check{\mathbf{R}}_I
\left\lfloor
\mathbf{J}\left(\mathbf{q}\right)\dot{\mathbf{q}}+\left(\boldsymbol{\omega}_t-\mathbf{b}^{\boldsymbol{\omega}}\times \mathbf{fk}\left(\mathbf{q}\right)\right)
\right\rfloor
_{\wedge} \\
\end{aligned}
\end{equation}
\subsubsection{{Position Measurement}}
The position residual measured from contact and forward kinematics can be written as: 
\begin{equation} ~\label{e: position residual z}
    \mathbf{z}_{cp} = {}^G\check{\mathbf{p}}_c - {}^G\check{\mathbf{p}}_I - {}^G\check{\mathbf{R}}_I \cdot \mathbf{fk}\left({\mathbf{q}}\right)
\end{equation}
Similarly, substituting (\ref{e: joint position measurement}) into (\ref{e: position residual z}) results in position measurement residual: 
\begin{equation} ~\label{e: position measurement h}
    \begin{aligned}
        \mathbf{0} &= \mathbf{h}_{cp}\left( \mathbf{x}_t, \mathbf{u}_t, \mathbf{w}^{\mathbf{q}}\right) \\
        &= \mathbf{h}_{cp}\left( \check{\mathbf{x}}_t, \mathbf{u}_t, \mathbf{0}\right) + \mathbf{H}_{cp}\tilde{\mathbf{x}}_t+ \mathbf{w}_{cp}\\
        &= \mathbf{z}_{cp} + \mathbf{H}_{cp}\tilde{\mathbf{x}}_t+ \mathbf{w}_{cp}
    \end{aligned}
\end{equation}
where $\mathbf{w}_{cp}\sim \mathcal{N}(\mathbf{0}, \mathbf{\Sigma}_{cp}) $ and
\begin{equation} ~\label{e: position H}
    \mathbf{H}_{cp} =
    \begin{bmatrix}
        {}^G\check{\mathbf{R}}_I \left\lfloor \mathbf{fk}\left(\mathbf{q}\right)\right\rfloor_{\wedge} 
        &
        -\mathbf{I}_3 &\mathbf{0}_{3\times 9} & \mathbf{I}_3 & \mathbf{0}_{3\times3}
    \end{bmatrix}
\end{equation}

\subsection{State Update}

Combining the prior in (\ref{e: imu model discretized}) with the measurement model from (\ref{e: lidar first taylor}), (\ref{e: kinematic first order taylor}), and (\ref{e: position measurement h}) yields the maximum a-posteriori (MAP) estimation: 
\begin{equation} ~\label{e: map eauqaiton}
\begin{aligned}
    \min _{\tilde{\mathbf{x}}_t}&\Bigg(\left\|\mathrm{x}_t \boxminus     \widehat{\mathbf{x}}_t\right\|_{\check{\mathbf{P}}_t^{-1}}^2
    +\sum_{j=1}^m\left\|\mathbf{h}_j+\mathbf{H}_j \tilde{\mathbf{x}}_t\right\|_{\mathbf{R}_j^{-1}}^2 \Bigg.+
    \\ &\left. \left\| \mathbf{h}_{cv}\left({\check{\mathbf{x}}}_t,\mathbf{u}_t,\mathbf{0}, \mathbf{0}\right) + \mathbf{H}_{cv }\tilde{\mathbf{x}}_t 
    \right\|_{\mathbf{\Sigma}_{cv}^{-1}}^2 \right. +
    \\  &\Bigg.\left\| \mathbf{h}_{cp}\left( \check{\mathbf{x}}_t, \mathbf{u}_t, \mathbf{0}\right) + \mathbf{H}_{cp}\tilde{\mathbf{x}}_t
    \right\|_{\mathbf{\Sigma}_{cp}^{-1}}^2
    \Bigg)
\end{aligned}
\end{equation}
Denote: 
\begin{equation} ~\label{e: kalman filter H, R, P}
    \begin{aligned}
        \mathbf{H} &= \begin{bmatrix}
            \mathbf{H}_1^T, \ldots , \mathbf{H}_j^T & \mathbf{H}_{cv} &\mathbf{H}_{cp}
        \end{bmatrix}^T \\
        \mathbf{R} &= \operatorname{diag}\left(\mathbf{R}_1,\ldots ,\mathbf{R}_j, \mathbf{\Sigma}_{{cv}}, \mathbf{\Sigma}_{{cp}}\right)\\
        \mathbf{z}_t &= 
        \begin{bmatrix}
            \mathbf{h}_1\left(\check{\mathbf{x}}_t,\mathbf{0}        \right),\ldots , \mathbf{h}_j\left(\check{\mathbf{x}}_t,\mathbf{0}\right) & \mathbf{h}_{cv}
            &
            \mathbf{h}_{cp}             
        \end{bmatrix}^T  \\
        \mathbf{P} &= \check{\mathbf{P}}_t 
    \end{aligned}
\end{equation}

Following \cite{xu2021fast}, the Kalman gain is computed as: 
\begin{equation}
\mathbf{K}=\left(\mathbf{H}^T \mathbf{R}^{-1} \mathbf{H}+\mathbf{P}^{-1}\right)^{-1} \mathbf{H}^T \mathbf{R}^{-1}  
\end{equation}
Then we can update the state estimate as: 
\begin{equation}
    \hat{\mathbf{x}}_t^{j+1}=\hat{\mathbf{x}}_t^j \boxplus\left(-\mathbf{K} {\mathbf{z}}_t^j-(\mathbf{I}-\mathbf{K} \mathbf{H})(\mathbf{J})^{-1}\left(\hat{\mathbf{x}}_t^j \boxminus \check{\mathbf{x}}_t\right)\right)
\end{equation}
The above process is iterated until convergence (i.e., the update is smaller than a given threshold). Then, the optimal estimation and covariance is: 
\begin{equation} ~\label{e: optimal estimation}
    \hat{\mathbf{x}}_t = \hat{\mathbf{x}}_t^{j+1}, \quad \hat{\mathbf{P}} = (\mathbf{I}-\mathbf{K} \mathbf{H}) \mathbf{P}
\end{equation}

\section{Experiments}

Inspired by FastLIO2 \cite{xu2022fast}, we achieved the C++ implementation of LIKO. The experiment and datasets were conducted on the BHR-B3 bipedal robot (shown in Fig. \ref{fig:teaser-fig}) using a walking controller from \cite{han2023heuristic}. The robot was controlled by an Intel NUC with AMD Ryzen 7 5700u and 16 GiB RAM. The sensor specifications of BHR-B3 are shown in Table \ref{tab: sensor spec}. A more detailed introduction of BHR-B3 is presented in \cite{gao2023hybrid}. The IMU, joint encoder, and F/T sensor are connected to the NUC via EtherCAT, while the LiDAR is connected using Ethernet. Additionally, we use Pinocchio \cite{pinocchio} for kinematic computation. 
\begin{table}[t]
    \centering
    \caption{Sensor Specifications of the BHR-B3 Robot}
    \begin{tabular}{llrl}
    \toprule
    Sensor               & Model                                                                     & \multicolumn{1}{c}{Hz} & Specs                                                                                 \\
    \midrule
    Encoder              & \begin{tabular}[c]{@{}l@{}}HEIDENHAIN \\ EBI-1135\end{tabular}            & 1000                   & \textit{Res: }$<0.0014^{\circ}$            \\
    \midrule
    
    F/T Sensor           & SRI M3714B4                                                               & 1000                   & \begin{tabular}[c]{@{}l@{}}\textit{Capacity: }\\ $x,y: 900\mathrm{N},100\mathrm{Nm}$  \\ $z: 1800\mathrm{N},100\mathrm{Nm}$\end{tabular} \\
    \midrule
    LiDAR                & \begin{tabular}[c]{@{}l@{}}Velodyne \\ VLP-16\end{tabular}                & 10                     & \textit{Res: }$16 \mathrm{px} \times 1824 \mathrm{px}$                                                    \\
    \midrule
    
    \multirow{2}{*}{IMU} & \multirow{2}{*}{\begin{tabular}[c]{@{}l@{}}Xsens \\ Mti-100\end{tabular}} & \multirow{2}{*}{200}   & \textit{Init Bias: }$0.2^{\circ} / \mathrm{s} \mid 5 \mathrm{mg}$                                                \\
    
                         &                                                                           &                        & \textit{Bias Stab: }$10^{\circ} / \mathrm{h} \mid 15 \mathrm{mg}$\\
    \bottomrule
    \end{tabular}
    \label{tab: sensor spec}
\end{table}

\subsection{Dataset}
  In this section, we introduce the dataset used to evaluate LIKO. The dataset was collected in a room equipped with a VICON MoCap system, where the BHR-B3 robot walked in different patterns. The LiDAR, IMU, F/T sensor, joint encoder, and VICON MoCap measurements are collected in the format of ROS bags \cite{quigley2009ros}. The VICON MoCap measurement is the ground truth in each dataset. Motion patterns performed in each dataset are listed as follows:
  \begin{itemize}
      \item \textit{forward\_backward: } The robot walks forward in a straight line and then reverses backward to the start point.
      \item \textit{square\_walk: }The robot walks along a square trajectory with a side length of about three meters, starting from the center of the square and returning to the center.
      \item \textit{square\_walk\_dynamic: } The robot follows the same trajectory as in \textit{square\_walk}, but returns to the center from another side. 
      \item \textit{walk\_in\_place: }The robot walks in place and rotates 90 degrees counterclockwise. 
      \item \textit{up\_slope: }The robot walks up a slope, stays at the top for a short time and then walks downward. 
  \end{itemize}

\begin{figure}[pt]
    \centering
    \includegraphics[width=0.85\linewidth]{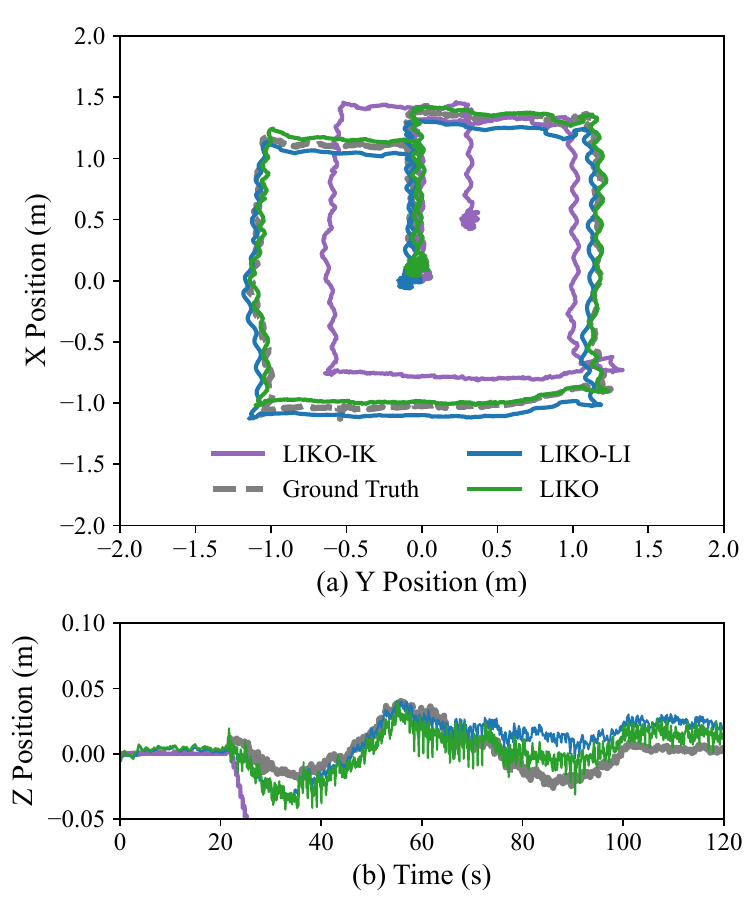}
    \caption{\textit{Top:} XY position trajectory estimation results of three LIKO variants used in the ablation study against ground truth (gray). \textit{Bottom: } The Z Position trajectory estimation results of three LIKO variants used in the ablation study against ground truth (gray).}
    \label{fig: ablation}
\end{figure}

\begin{table*}[pt]
    \centering
        \caption{Accuracy Evaluation (RMSE) in Sequences With Ground Truth}
\begin{tabular}{lccccccccccc}
\toprule
        \multirow{2}{*}{Algorithm} & \multicolumn{2}{c}{\textit{forward\_backward}}        & \multicolumn{2}{c}{\textit{square\_walk}}        & \multicolumn{2}{c}{\textit{square\_walk\_long}}        & \multicolumn{2}{c}{\textit{walk\_in\_place}}        & \multicolumn{2}{c}{\textit{up\_slope}}        & Frequency \\
         \cmidrule(l){2-3} \cmidrule(l){4-5} \cmidrule(l){6-7} \cmidrule(l){8-9} \cmidrule(l){10-11}
         & APE (m)         & RPE(\%)         & APE (m)         & RPE(\%)         & APE (m)         & RPE(\%)         & APE (m)         & RPE(\%)         & APE (m)         & RPE(\%)         & (Hz)      \\
         \midrule
FastLIO2 & 0.0117          & 0.0052          & 0.0294          & 0.0059          & 0.0200          & 0.0067          & 0.0341          & 0.0058          & 0.0299          & 0.0054          & 10        \\
LIO-SAM  & 0.0180          & 0.0035          & 0.0305          & 0.0037          & 0.0189          & 0.0037          & \textbf{0.0167}          & 0.0036          & 1.7142          & 0.0265          & 1000      \\
LINS     & 0.0209          & 0.0239          & 0.0231          & 0.0257          & 0.0245          & 0.0270          & 0.0204 & 0.0227          & 1.6817          & 0.0528          & 3         \\
LIKO (Ours)     & \textbf{0.0091} & \textbf{0.0006} & \textbf{0.0198} & \textbf{0.0018} & \textbf{0.0180} & \textbf{0.0021} & 0.0215          & \textbf{0.0011} & \textbf{0.0245} & \textbf{0.0013} & 1000    \\ 
\bottomrule
\label{t: main table}
\end{tabular}
\end{table*}

\begin{figure}[pt]
    \centering
    \includegraphics[width=1\linewidth]{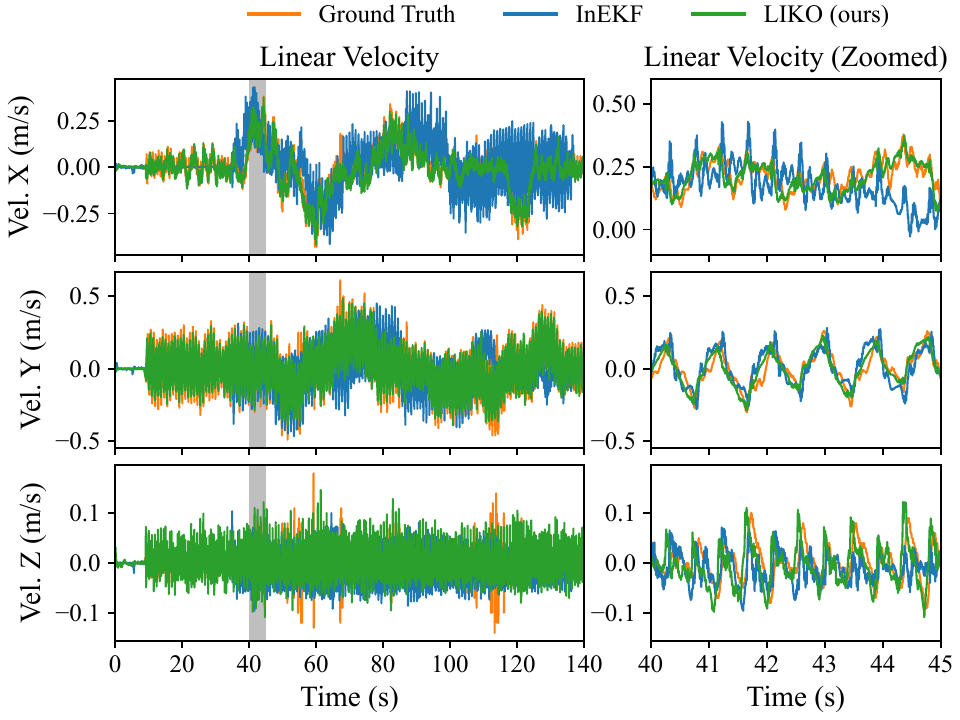}
    \caption{\textit{Left:} Linear velocity comparison between LIKO (green) and Contact-aided InEKF \cite{hartley2020contact} (blue) against ground truth (orange). \textit{Right:} Zoomed-in plot corresponding to the gray area of the left one. The more accurate and smoother velocity tracking allows for a better closed-loop control performance.}
    \label{fig:cmp_riekf_liko}
\end{figure}

\subsection{Ablation Study}
In this section, we study the individual contribution of different sensor modalities. The set of combinations tested are as follows: 1) LIKO-IK: IMU and leg odometry are used. In this way, LIKO-IK is a proprioceptive odometry that uses IEKF for sensor fusion. 2) LIKO-LI: LiDAR and IMU are used. This is similar to FastLIO2 but the odometry output frequency is 1kHz. 3) LIKO: uses LiDAR, IMU, and leg odometry. 

We used the \textit{square\_walk} dataset for the ablation study, the result is shown in Fig. \ref{fig: ablation}. Note that LIKO-LI has performance closer to LIKO than LIKO-IK. This suggests that the LiDAR sensor plays the main contribution in global position estimation. Although LIKO-IK drifts in position estimation, the high-frequency nature of leg odometry allows it to provide velocity measurements more timely and accurately than LIO methods. 

\subsection{Comparison With Other Algorithms}
\subsubsection{Comparison with LIO algorithms}
This section compares the LIKO against state-of-the-art LiDAR-Inertial mapping systems, including FastLIO2 \cite{xu2022fast}, LIO-SAM \cite{liosam2020shan} and LINS \cite{qin2020lins}. We use the absolute pose error (APE) as an accuracy indicator for whole trajectories and a translational relative pose error (RPE) for drift evaluation. Both APE and RPE are evaluated using evo \cite{grupp2017evo}. 

The accuracy result is shown in Table \ref{t: main table}. We achieve state-of-the-art evaluation accuracy in four of the five datasets. The only exception is on \textit{walk\_in\_place}, where LIO-SAM and LINS show slightly higher accuracy than LIKO in APE. In the \textit{up\_slope} dataset, LIO-SAM and LINS showed a great drift, while LIKO remains high accuracy. The reason is that the robot traversed uneven terrain in this dataset, and the leg odometry played a key role in such circumstances.

\subsubsection{Comparison with inertial-kinematic algorithms}
This section compares the LIKO against state-of-art Kinematic-Inertial odometry, contact-aided InEKF \cite{hartley2020contact}. Since the position is unobservable for proprioceptive odometry, we only compare the linear velocity estimation of two algorithms, the result is shown in Fig. \ref{fig:cmp_riekf_liko}. The velocity ground truth was obtained by taking the time derivative of the MoCap data. A 20-data-wide sliding window filter was applied to the ground truth velocity to remove extra noise due to MoCap marker loss. It can be seen that LIKO achieves more accurate and smooth velocity estimation than contact-aided InEKF in the x and y directions. The robot's velocity estimation in the z-direction is similar to that of the other two directions due to walking on flat ground with less disturbance. 

\subsection{Contact Position Estimation}
\begin{figure}[pt]
    \centering
    \includegraphics[width=0.96\linewidth]{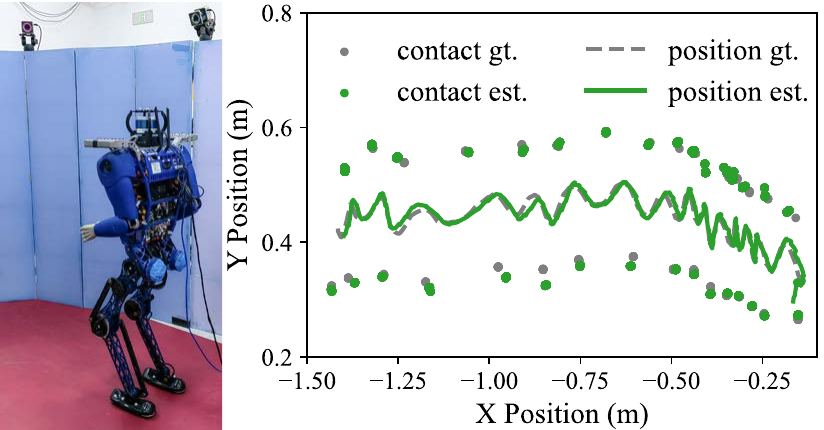}
    \caption{\textit{Left: } BHR-B3 walking in the MoCap room. Head and feet are equipped with MoCap markers. \textit{Right: }Top-down view of the trajectory and contact position estimation (green) and the ground truth (gray). In the legend, "gt." stands for "ground truth" and "est." stands for "estimated".}
    \label{fig:contact_position}
\end{figure}
The foot contact position estimation was also evaluated with MoCap Ground Truth, as shown in Fig. \ref{fig:contact_position}. MoCap markers are attached to the feet of the robot to measure contact position during locomotion. It can be seen that the LIKO provides an accurate estimation of the foot contact position. Note that the contact point in the graph is enlarged for better visualization and each contact position contains all of the points in its phase. This estimation can be used as feedback for the foothold plan and control module, providing a more robust locomotion behavior.

\section{Conclusions}
We have presented LIKO, a tightly coupled LiDAR-inertial-kinematic biped robot state estimation algorithm with online foothold estimation. Extensive experiments on the biped robot platform have shown that our algorithm outperforms many existing LiDAR-inertial or inertial-kinematic state estimation methods. The source code and dataset will be open-sourced to benefit the community. 

We believe that LIKO's accuracy will further improve with more detailed modeling of flat foot contact and kinematic parameters \cite{yang2022online} addressed. In the future, we plan to develop corresponding downstream control applications and introduce visual sensors to help our robot conquer more challenging environments.






\bibliography{ref}

\end{document}